\documentclass[runningheads]{llncs}

\usepackage[mobile]{eccv}

\usepackage{eccvabbrv}

\usepackage{graphicx}
\usepackage{booktabs}

\usepackage[accsupp]{axessibility}  

\usepackage{amsmath}
\usepackage{amssymb}
\usepackage{cases}
\usepackage{algorithm}
\usepackage{algpseudocode}
\usepackage{xcolor}
\usepackage{overpic}
\usepackage{rotating}
\usepackage[accsupp]{axessibility} 
\usepackage{multirow}
\usepackage{subcaption}

\usepackage{hyperref}

\usepackage{orcidlink}

\usepackage[capitalize]{cleveref}
\usepackage{wrapfig}
\crefname{section}{Sec.}{Secs.}
\Crefname{section}{Section}{Sections}
\Crefname{table}{Table}{Tables}
\crefname{table}{Tab.}{Tabs.}
\Crefname{subfigure}{Figure}{Figures}
\crefname{subfigure}{Fig.}{Figs.}

\begin{document}

\newcommand{\method}{SlimFlow}
\newcommand{\anflow}{Annealing Reflow}

\title{SlimFlow: Training Smaller One-Step Diffusion Models with Rectified Flow}


\author{Yuanzhi Zhu\inst{1}
\and
Xingchao Liu\inst{2} \and
Qiang Liu\inst{2} }

\authorrunning{Zhu Y., Liu X., Liu Q.}

\institute{ETH Zurich, Zurich 8092, Switzerland \\
\email{yuazhu@student.ethz.ch} 
\and
UT Austin, Austin TX 78712, United States\\
\email{xcliu@utexas.edu} \quad \email{lqiang@cs.utexas.edu}}
\maketitle

\begin{abstract}
  Diffusion models excel in high-quality generation but suffer from slow inference due to iterative sampling. While recent methods have successfully transformed diffusion models into one-step generators, they neglect model size reduction, limiting their applicability in compute-constrained scenarios. 
  This paper aims to develop small, efficient one-step diffusion models based on the powerful rectified flow framework, by exploring joint compression of inference steps and model size.
  The rectified flow framework trains one-step generative models using two operations, reflow and distillation.
  Compared with the original framework, squeezing the model size brings two new challenges: 
  (1) the initialization mismatch between large teachers and small students during reflow;
  (2) the underperformance of naive distillation on small student models.
  To overcome these issues, we propose Annealing Reflow and Flow-Guided Distillation, which together comprise our \method{} framework.
  With our novel framework, we train a one-step diffusion model with an FID of 5.02 and 15.7M parameters, outperforming the previous state-of-the-art one-step diffusion model (FID=6.47, 19.4M parameters)
  on CIFAR10.
  On ImageNet 64$\times$64 and FFHQ 64$\times$64, our method yields small one-step diffusion models that are comparable to larger models, showcasing the effectiveness of our method in creating compact, efficient one-step diffusion models.
  
\end{abstract}

\section{Introduction}
\label{sec:intro}

In recent years, diffusion models~\cite{song2020score, ho2020denoising} have revolutionized the field of image generation~\cite{ramesh2022hierarchical,nichol2021improved,dhariwal2021diffusion, liu2022let}, surpassing the quality achieved by traditional approaches such as Generative Adversarial Networks (GANs)~\cite{goodfellow2020generative,karras2019style}, Normalizing Flows (NFs)~\cite{dinh2016density, papamakarios2021normalizing} and Variational Autoencoders (VAEs)~\cite{kingma2013auto,razavi2019generating}. Its success even extends to other modalities, including audio~\cite{kong2020diffwave,jeong2021diff}, video~\cite{ho2022imagen,molad2023dreamix,geyer2023tokenflow} and 3D content generation~\cite{poole2022dreamfusion, wang2023score, liu2023zero, wu2022diffusion}. 

Despite these advancements, diffusion models face significant challenges in terms of generation efficiency. Their iterative sampling process and substantial model size pose considerable obstacles to widespread adoption, particularly in resource-constrained environments, such as edge devices.
To enhance the inference speed of diffusion models, recent research has focused on two primary strategies.
The first strategy aims at reducing the number of inference steps, which can be achieved through either adoption of fast solvers~\cite{song2020denoising,lu2022dpm,zhang2022fast,dockhorn2022genie,zhou2023fast} or distillation~\cite{luhman2021knowledge,salimans2022progressive,gu2023boot,song2023consistency, liu2022rectified}.
The second strategy focuses on lowering the computational cost within each inference step, by model structure modification~\cite{fang2024structural, yang2023diffusion,pernias2023wurstchen,crowson2024scalable}, quantization~\cite{wang2023towards, li2023q, he2023efficientdm, shang2023post, li2024q, huang2023tfmq}, and caching~\cite{ma2023deepcache,wimbauer2023cache}.
Recently, these techniques have been applied to various large-scale diffusion models, e.g., Stable Diffusion~\cite{rombach2022high}, to significantly improve their inference speed and reduce their cost~\cite{meng2023distillation, luo2023latent, liu2023instaflow, yin2024one, xu2023ufogen}.

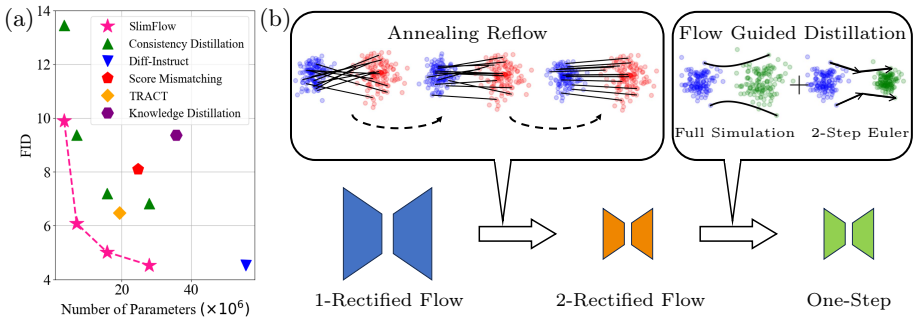
\begin{figure}[!t]
\centering
\begin{overpic}[width=1.\linewidth]{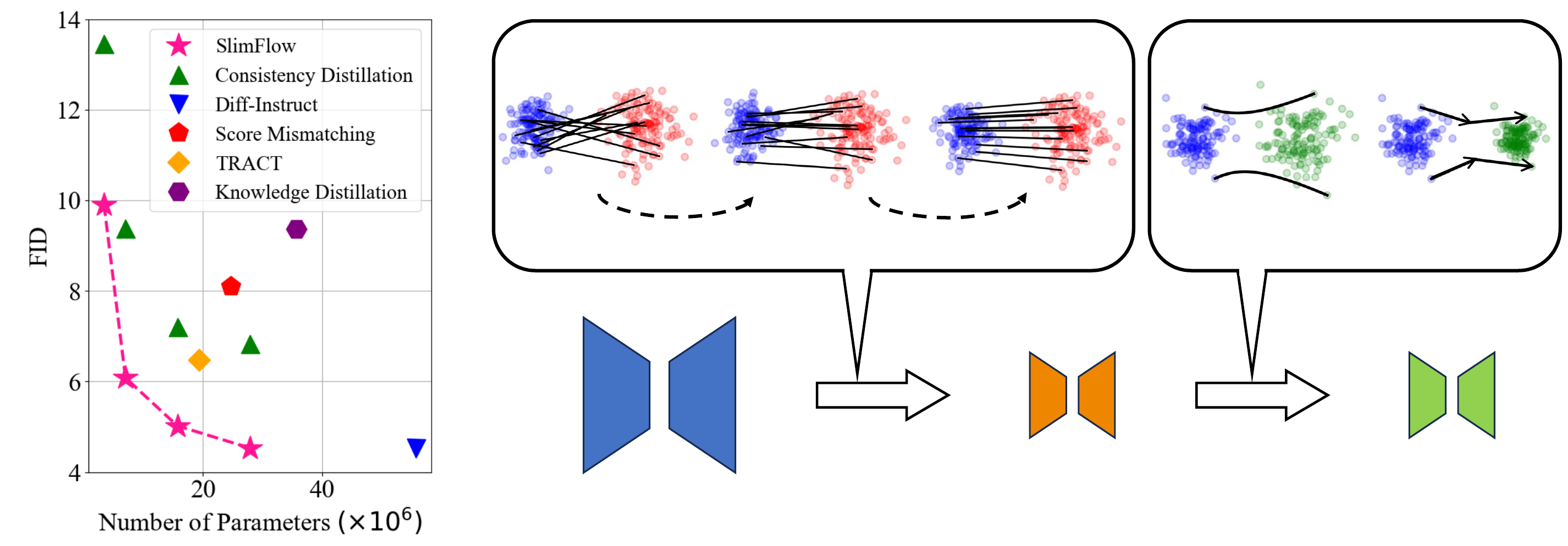}
\put(0,32){\color{black}{(a)}}
\put(28.,32){\color{black}{(b)}}
\put(86.,25){\color{black}\footnotesize{\textbf{+}}}
\put(42,30.6){\color{black}\scriptsize{Annealing Reflow}}
\put(74,30.6){\color{black}\scriptsize{Flow Guided Distillation}}
\put(73.5,20){\color{black}\tiny{Full Simulation}}
\put(88.5,20){\color{black}\tiny{2-Step Euler
}}
\put(34,1.8){\color{black}\scriptsize{1-Rectified Flow}}
\put(60.5,1.8){\color{black}\scriptsize{2-Rectified Flow}}
\put(88,1.8){\color{black}\scriptsize{One-Step}}
\end{overpic}
\caption{
(a) Comparison of different one-step diffusion models on the CIFAR10 dataset.
(b) To get powerful one-step diffusion model, our \method{} framework designs two stages: 
\textbf{Annealing Reflow} provides a warm-start for the small 2-Rectified Flow model by gradually shifting from training with random pairs to teacher pairs; 
\textbf{Flow Guided Distillation} enhances the one-step small model by distillation from 2-Rectified Flow with both off-line generated data using precise ODE solver and online generated data using 2-step Euler solver.
}
\label{fig:main}
\end{figure}

This paper focuses on advancing the rectified flow framework \cite{liu2022flow, liu2022rectified}, which has demonstrated promising results in training few/one-step diffusion models \cite{liu2023instaflow}. While the framework has shown success in reducing inference steps, it has not addressed the challenge of model size reduction. We aim to enhance the rectified flow framework by simultaneously reducing both the number of inference steps and the network size of diffusion models.
The rectified flow framework straightens the trajectories of pre-trained generative probability flows through a process called reflow, thereby decreasing the required number of inference steps. This procedure also refines the coupling between noise and data distributions. High-quality one-step generative models are then obtained by distilling from the straightened flow.
Unlike typical rectified flow applications that maintain an invariant model structure during the entire process, we aim to reduce the network size in reflow and distillation, targeting efficient one-step diffusion models.
It introduces two key challenges: 
(1) The reflow operation typically initializes the student flow with the pre-trained teacher's weights to inherit knowledge and accelerate convergence. However, this strategy is inapplicable when the student network has a different structure.
(2) The smaller student network suffers from reduced capacity, causing naive distillation to underperform.
To address these challenges, we propose~\method{}, comprising two stages: Annealing Reflow and Flow-Guided Distillation.
Annealing Reflow provides a warm-start initialization for the small student model by smoothly interpolating between training from scratch and reflow. 
Flow-Guided Distillation introduces a novel regularization that leverages guidance from the learned straighter student flow, resulting in better distilled one-step generators.
\method{} achieves state-of-the-art Frechet Inception Distance (FID) among other one-step diffusion models with a limited number of parameters.
Furthermore, when applied to ImageNet 64$\times$64 and FFHQ 64$\times$64, \method{} yields small one-step diffusion models that are comparable to larger models.

\section{Background}
\subsection{Diffusion Models}
Diffusion models define a forward diffusion process that maps data to noise by gradually perturbing the input data with Gaussian noise.
Then, in the reverse process, they generate images by gradually removing Gaussian noise, with the intuition from non-equilibrium thermodynamics \cite{sohl2015deep}. 
We denote the data $\mathbf{x}$ at $t$ as $\mathbf{x}_t$.
The forward process can be described by an It\^{o} SDE \cite{song2020score}:
\begin{equation}\label{eq:itoSDE}
    \mathrm{d}\mathbf{x}_t = \mathbf{f}(\mathbf{x}_t, t) \mathrm{d}t + g(t) \mathrm{d} \mathbf{w},
\end{equation}
where $\mathbf{w}$ is the standard Wiener process, $\mathbf{f}(\cdot,t)$ is a vector-valued function called the drift coefficient, and ${g}(\cdot)$ is a scalar function called the diffusion coefficient. 

For every diffusion process in \cref{eq:itoSDE}, there exists a corresponding deterministic Probability Flow Ordinary Differential Equation (PF-ODE) which induces the same marginal density as~\cref{eq:itoSDE}:
\begin{equation}\label{eq:pf_ode}
\frac{\mathrm{d}\mathbf{x}_t}{\mathrm{d}t} = \mathbf{f}(\mathbf{x}_t, t) - \frac{1}{2} g^2(t) \nabla_{\mathbf{x}_t} \log p_t(\mathbf{x}_t),
\end{equation}
where $p_t(\cdot)$ is the marginal probability density at time $t$. 
$\nabla_{\mathbf{x}_t} \log p_t(\mathbf{x}_t)$ is called the score function, and can be modelled as $\mathbf{s}_\theta(\mathbf{x},t)$ using a neural network $\theta$. Usually, the network is trained by score matching~\cite{hyvarinen2005estimation,song2019generative,song2020sliced}.
Starting with samples from an initial distribution $\pi_T$ such as a standard Gaussian distribution, we can generate data samples by simulating \cref{eq:pf_ode} from $t=T$ to $t=0$. We call \cref{eq:pf_ode} with the approximated score function $\mathbf{s}_\theta(\mathbf{x}_t,t)$ the empirical PF-ODE, written as $\frac{\mathrm{d}\mathbf{x}_t}{\mathrm{d}t} = \mathbf{f}(\mathbf{x}_t, t) - \frac{1}{2} g^2(t) \mathbf{s}_\theta(\mathbf{x}_t,t)$.
It is worth noting that the deterministic PF-ODE gives a deterministic correspondence between the initial noise distribution and the generated data distribution.

\subsection{Rectified Flows}
Rectified flow~\cite{liu2022flow, liu2022rectified, lipman2022flow} is an ODE-based generative modeling framework. Given the initial distribution $\pi_T$ and the target data distribution $\pi_0$, rectified flow trains a velocity field parameterized by a neural network with the following loss function,
\begin{equation}\label{eq:objective_rf}
\begin{aligned}
   \mathcal{L}_{\text{rf}}(\theta):=\mathbb{E}_{\mathbf{x}_T \sim \pi_T, \mathbf{x}_0 \sim \pi_0}
   &\left[ \int_0^T \big \| \mathbf{v}_\theta (\mathbf{x}_t, t ) - (\mathbf{x}_T - \mathbf{x}_0) \big \|_2^2 \mathrm{d}t \right],\\
   \text{where}~~~~&\mathbf{x}_t = (1-t/T)\mathbf{x}_0 + t\mathbf{x}_T/T.
\end{aligned}
\end{equation}
Without loss of generalization, $T$ is usually set to $1$. Based on the trained rectified flow, we can generate samples by simulating the following ODE from $t=1$ to $t=0$,
\begin{equation}
\label{eq:rf_ode}
    \frac{\mathrm{d}\mathbf{x}_t}{\mathrm{d}t} = \mathbf{v}_\theta(\mathbf{x}_t, t).
\end{equation}
PF-ODEs transformed from pre-trained diffusion models can be seen as special forms of rectified flows. For a detailed discussion of the mathematical relationship between them, we refer readers to~\cite{liu2022flow, liu2022rectified}. In computer, ~\cref{eq:rf_ode} is approximated by off-the-shelf ODE solvers, e.g., the forward Euler solver,
\begin{equation}
\label{eq:rf_ode_discrete}
    \mathbf{x}_{t - \frac{1}{N}} = \mathbf{x}_{t} - \frac{1}{N} \mathbf{v}_\theta(\mathbf{x}_{t}, t),~~~\forall t \in \{1, 2, \dots, N \} / N. 
\end{equation}
Here, the ODE is solved in $N$ steps with a step size of $1/N$. Large $N$ leads to accurate but slow simulation, while small $N$ gives fast but inaccurate simulation. 
Fortunately, straight probability flows with uniform speed enjoy one-step simulation with no numerical error because $\mathbf{x}_{t} = \mathbf{x}_{1} - (1-t)\mathbf{v}_\theta(\mathbf{x}_{1}, 1)$.

\noindent\textbf{Reflow}~~
In the rectified flow framework, a special operation called reflow is designed to train straight probability flows,
\begin{equation}\label{eq:objective_reflow}
\begin{aligned}
   \mathcal{L}_{\text{reflow}}(\phi)&:=\mathbb{E}_{\mathbf{x}_1 \sim \pi_1}
   \left[ \int_0^T \big \| \mathbf{v}_\phi (\mathbf{x}_t, t ) - (\mathbf{x}_1 - \hat{\mathbf{x}}_0) \big \|_2^2 \mathrm{d}t \right],\\
   \text{where}~~~~&\mathbf{x}_t = (1-t)\hat{\mathbf{x}}_0 + t\mathbf{x}_1~~\text{and}~~\hat{\mathbf{x}}_0 = \mathrm{ODE}[\mathbf{v}_\theta](\mathbf{x}_1).
\end{aligned}
\end{equation}
Compared with~\cref{eq:objective_rf}, $\hat{\mathbf{x}}_0$ is not a random sample from distribution $\pi_0$ that is independent with $\mathbf{x}_1$  anymore. Instead, $\hat{\mathbf{x}}_0 = \mathrm{ODE}[\mathbf{v}_\phi](\mathbf{x}_1) := \mathbf{x}_1 + \int_1^0 \mathbf{v}_\theta(\mathbf{x}_{t},t) \mathrm{d} t$ is induced from the pre-trained flow $\mathbf{v}_\theta$. After training, the new flow $\mathbf{v}_\phi$ has straighter trajectories and requires fewer inference steps for generation. 
A common practice in reflow is initializing $\mathbf{v}_\phi$ with the pre-trained $\mathbf{v}_\theta$ for faster convergence and higher performance~\cite{liu2022flow}. 
Following previous works, we name $\mathbf{v}_\theta$ as 1-rectified flow and $\mathbf{v}_\phi$ as 2-rectified flow. 
The straightness of the learned flow $\mathbf{v}$ can be defined as:
\begin{equation}\label{eq:straightness}
\begin{aligned}
   S(\mathbf{v}) = \int_0^1 \mathbb{E}\left[\| 
\mathbf{v}(\mathbf{x}_t, t) - (\mathbf{x}_1 - \mathbf{x}_0)  \|^2\right] \mathrm{d} t,
\end{aligned}
\end{equation}

\noindent\textbf{Distillation}~~
Given a pre-trained probability flow, for example, $\mathbf{v}_\phi$, we can further enhance their one-step generation via distillation,
\begin{equation}
\label{eq:distill}
    \mathcal{L}_{\text{distill}}(\phi'):= \mathbb{E}_{\mathbf{x}_1 \sim \pi_1} \left [ \mathbb{D}(\mathrm{ODE}[\mathbf{v}_\phi](\mathbf{x}_1), \mathbf{x}_1 - \mathbf{v}_{\phi'}(\mathbf{x}_1, 1)) \right].
\end{equation}
In this equation, $\mathbb{D}$ is a discrepancy loss that measures the difference between two images, e.g., $\ell_2$ loss or the LPIPS loss~\cite{zhang2018perceptual}. Through distillation, the distilled model $\mathbf{v}_{\phi'}$ can use one-step Euler discretization to approximate the result of the entire flow trajectory. 
Note that reflow can refine the deterministic mapping between the noise and the generated samples defined by the probability flow~\cite{liu2022flow, liu2023instaflow}. 
Consequently, distillation from $\mathbf{v}_\phi$ gives better one-step generators than distillation from $\mathbf{v}_\theta$.

\section{\method{}}
\label{sec:method}

In this section, we introduce the \method{} framework for learning small, efficient one-step generative models.
\method{} enhances the rectified flow framework by improving both the reflow and distillation stages.
For the reflow stage, we propose \anflow{}. This novel technique provides a warm-start initialization for the small student model. It smoothly transitions from training a 1-rectified flow to training a 2-rectified flow, accelerating the training convergence of the smaller network.
For the distillation stage, we propose Flow-Guided Distillation, which leverages the pre-trained 2-rectified flow as an additional regularization during distillation to improve the performance of the resulting one-step model.
By combining these two techniques, \method{} creates a robust framework for training efficient one-step diffusion models that outperform existing methods in both model size and generation quality.

\subsection{\anflow{}}
\label{sec:an_flow}

The reflow procedure in the rectified flow framework trains a new, straighter flow $\mathbf{v}_\phi$ from the pre-trained 1-rectified flow $\mathbf{v}_\theta$, as described in ~\cref{eq:objective_reflow}. Typically, when $\mathbf{v}_\phi$ and $\mathbf{v}_\theta$ have the same model structure, $\mathbf{v}_\phi$ is initialized with $\mathbf{v}_\theta$'s weights to accelerate training. However, our goal of creating smaller models cannot adopt this approach, as $\mathbf{v}_\phi$ and $\mathbf{v}_\theta$ have different structures.
A naive solution would be to train a new 1-rectified flow using the smaller model with~\cref{eq:objective_rf} and use this as initialization. However, this strategy is time-consuming and delays the process of obtaining an efficient 2-rectified flow with the small model.

To address this challenge, we propose \anflow{}, which provides an effective initialization for the small student model without significantly increasing training time. This method starts the training process with random pairs, as in~\cref{eq:objective_rf}, then gradually shifts to pairs generated from the teacher flow, as in~\cref{eq:objective_reflow}.

Formally, given the teacher velocity field $\mathbf{v}_\theta$ defined by a large neural network $\theta$, the objective of \anflow{} is defined as,
\begin{equation}
\label{eq:objective_an_reflow}
\begin{aligned}
   \mathcal{L}^{k}_{\text{a-reflow}}(\phi):=&\mathbb{E}_{\mathbf{x}_1, \mathbf{x}'_1 \sim \pi_1} 
   \left[ \int_0^T \big \| \mathbf{v}_\phi (\mathbf{x}_t^{\beta(k)}, t ) - \left(\mathbf{x}_1^{\beta(k)} - \hat{\mathbf{x}}_0 \right ) \big \|_2^2 \mathrm{d}t \right], \\
   \text{where}~~~~& \mathbf{x}_t^{\beta(k)} = (1-t) \hat{\mathbf{x}}_0 + t \mathbf{x}_1^{\beta(k)}, \\ &\mathbf{x}_1^{\beta(k)} = \left ( \sqrt{1-\beta^2(k)} \mathbf{x}_1 + \beta(k) \mathbf{x}'_1  \right ), \\
   &\hat{\mathbf{x}}_0 = \mathrm{ODE}[\mathbf{v}_\theta](\mathbf{x}_1) = \mathbf{x}_1 + \int_1^0 \mathbf{v}_\theta(\mathbf{x}_{t},t) \mathrm{d} t.
\end{aligned}
\end{equation}
In~\cref{eq:objective_an_reflow}, $\mathbf{x}_1$ and $\mathbf{x}'_1$ are independent random samples from the initial distribution $\pi_1$. $k$ refers to the number of training iterations, and $\beta(k):\mathbb{N} \rightarrow [0,1]$ is a function that maps $k$ to a scalar between $0$ and $1$. The function $\beta(k)$ must satisfy two key conditions:
\begin{enumerate}
    \item $\beta(0)=1$: This initial condition reduces $\mathcal{L}^{0}_{\text{a-reflow}}$ to the rectified flow objective (\cref{eq:objective_rf}), with $\hat{\mathbf{x}}_0$ generated from the well-trained teacher model instead of being randomly sampled from $\pi_0$. Note that when $\mathbf{v}_\theta$ is well-trained, $\hat{\mathbf{x}}_0$ also follows the target data distribution $\pi_0$.
    \item $\beta(+\infty)=0$: This asymptotic condition ensures that $\mathcal{L}^{+\infty}_{\text{a-reflow}}$ converges to the reflow objective (\cref{eq:objective_reflow}), guaranteeing that the small model will eventually become a valid 2-rectified flow.
\end{enumerate}
By carefully choosing the schedule $\beta$, we create a smooth transition from training the small student model using random pairs to using pairs generated from the large teacher flow. This approach provides an appropriate initialization for the student model and directly outputs a small 2-rectified flow model, balancing efficiency and effectiveness in the training process.

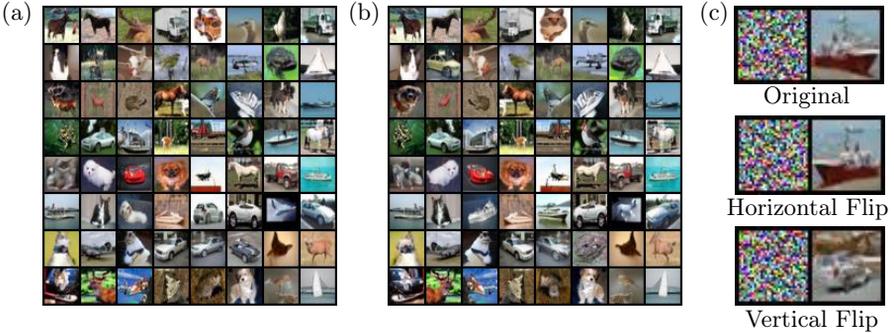
\begin{figure}[!t]
\centering
\begin{overpic}[width=1.\linewidth]{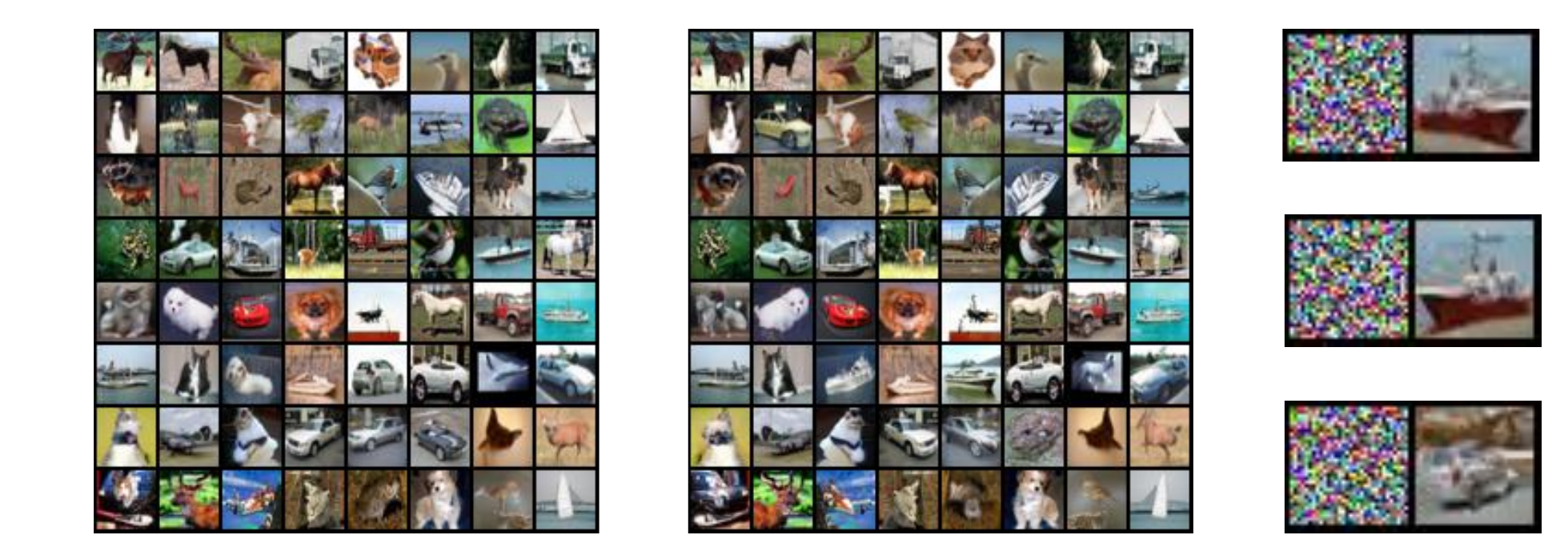}
\put(1.5,32){\color{black}{(a)}}
\put(39.5,32){\color{black}{(b)}}
\put(78,32){\color{black}{(c)}}
\put(85,23){\color{black}\footnotesize{Original}}
\put(81,11){\color{black}\footnotesize{Horizontal Flip}}
\put(83,-1.){\color{black}\footnotesize{Vertical Flip}}
\end{overpic}
\caption{
(a) Generation from 1-Rectified Flow trained without data augmentation.
(b) Generation from the 1-Rectified Flow model in (a) after applying horizontal flip to the same set of random noises in (a). 
(c) Horizontally flipping the noise results in horizontally flipped generated image, but vertical flip does not result in vertically flipped generated image.
}
\label{fig:symmetry}
\end{figure}

\noindent\textbf{Leveraging the Intrinsic Symmetry in Reflow Pairs.}~~
To perform reflow, a dataset of pair $(\mathbf{x}_1, \hat{\mathbf{x}}_0)$ is usually generated by the teacher flow before training the student model.
We find that the hidden symmetry in the pairs can be exploited to create new pairs without simulating the teacher flow.
Specifically, if the initial noise $\mathbf{x}_1$ is horizontally flipped (referred to as $\text{H-Flip}$), the corresponding image $\hat{\mathbf{x}}_0$ is also horizontally flipped.
As a result, for each generated pair $(\mathbf{x}_1, \hat{\mathbf{x}}_0)$, we can also add $\left( \text{H-Flip}(\mathbf{x}_1), \text{H-Flip}(\hat{\mathbf{x}}_0) \right )$ as a valid pair to double the sample size of our reflow dataset.


\subsection{Flow-Guided Distillation}

Distillation is the other vital step in training one-step diffusion models with the rectified flow framework.
To distill the student flow, naive distillation~(\cref{eq:distill}) requires simulating the entire student 2-rectified flow with off-the-shelf solvers, e.g., Runge-Kutta 45 (RK45) solver, to get the distillation target $\mathrm{ODE}[\mathbf{v}](\mathbf{x}_1)$. 
In most times, practitioners generate a dataset of $M$ teacher samples in advance to training the student model, denoted as $\mathcal{D}_{distill} = \left\{\left(\mathbf{x}_1^{(i)}, \mathrm{ODE}[\mathbf{v}](\mathbf{x}_1^{(i)})\right)\right\}_{i=1}^M.$
Such simulation takes tens of steps,
leading to excessive time and storage for the distilled one-step model. 
On the one hand, for small models in our scenario, we want to use large $M$ to reach satisfying one-step models due to their limited capacity.
On the other hand, we also expect to avoid costly data generation in the distillation stage.

To balance this trade-off, we propose flow-guided distillation, where we complement the expensive direct distillation with an additional regularization. This regularization is based on few-step generation from the 2-rectified flow $\mathbf{v}_\phi$. When estimating the integration result of $\mathrm{ODE}[\mathbf{v}_\phi](\mathbf{x}_1)$, we can adopt the forward Euler solver,
\begin{equation}
    \mathbf{x}_{t_i} = \mathbf{x}_{t_{i+1}} + (t_{i} - t_{i+1}) \mathbf{v}_\phi(\mathbf{x}_{t_{i+1}}, t_{i+1}),~~~\forall i \in \{0, 1, \dots, N-1 \},
\end{equation}
where the $N$ time steps $\{t_i\}_{i=0}^N$ are defined by the user and $t_{0}=0, t_{N}=1$. 
To find an appropriate supervision for one-step simulation without cumbersome precise simulation with advanced solvers, a feasible choice is two-step simulation, which is,
\begin{equation}
\label{eq:euler_2_step}
\begin{aligned}
    \hat{\mathbf{x}}_0 = \mathbf{x}_1 - (1-t) \mathbf{v}_\phi(\mathbf{x}_1, 1) - t\mathbf{v}_\phi(\mathbf{x}_t, t).
\end{aligned}
\end{equation}
Here, $t$ can be an intermediate time point between $(0, 1)$. Using this two-step approximation, we get another distillation loss,

\begin{equation}
\label{eq:2-step}
    \mathcal{L}_{\text{2-step}}(\phi'):= \mathbb{E}_{\mathbf{x}_1 \sim \pi_1} \left [ \int_0^1 \mathbb{D}(\mathbf{x}_1 - (1-t) \mathbf{v}_\phi(\mathbf{x}_1, 1) - t\mathbf{v}_\phi(\mathbf{x}_t, t), \mathbf{x}_1 - \mathbf{v}_{\phi'}(\mathbf{x}_1, 1)) \text{d} t \right].
\end{equation}
While this two-step generation is not as accurate as the precise generation $\mathrm{ODE}[\mathbf{v}_\phi](\mathbf{x}_1)$ with Runge-Kutta solvers, it can serve as a powerful additional regularization in the distillation process to boost the performance of the student one-step model. Because two-step generation is fast, we can compute this term online as extra supervision to the student one-step model without the need to increase the size of $\mathcal{D}_{distill}$. Our final distillation loss is,
\begin{equation}
    \mathcal{L}_{\text{combined}}(\phi') = \mathcal{L}_{\text{distill}}(\phi') + \mathcal{L}_{\text{2-step}}(\phi').
\end{equation}
It is worth mentioning that we can use more simulation steps to improve the accuracy of the the few-step regularization within the computational budget. In our experiments, using two-step generation as our regularization already gives satisfying improvement. We leave the discovery of other efficient regularization as our future work. The whole procedure of our distillation stage is captured in Algorithm~\ref{alg:Distillation}.

\begin{algorithm}[t]
    \small
    \caption{Flow-Guided Distillation}
    \label{alg:Distillation}
    \begin{algorithmic}[1]
    \Require Pre-trained 2-rectified flow $v_\phi$, dataset $\mathcal{D}_{distill}$ generated with $v_\phi$.
    \State{Initialize the one-step student model $\mathbf{v}_{\phi'}$ with the weights of $\mathbf{v}_{\phi}$.}
    \Repeat
        \State{Randomly sample $\left(\mathbf{x}_1, \mathrm{ODE}[\mathbf{v}_\phi](\mathbf{x}_1) \right) \sim \mathcal{D}_{distill}$.}
        \State{Compute $\mathcal{L}_{\text{distill}}(\phi')$ with~\cref{eq:distill}.}

        \State{Randomly sample $\mathbf{x}_1 \sim \pi_1$.}
        \State{Compute $\mathcal{L}_{\text{2-step}}(\phi')$ with~\cref{eq:2-step}.}

        \State{Compute $\mathcal{L}_{\text{combined}}(\phi')=\mathcal{L}_{\text{distill}}(\phi')+\mathcal{L}_{\text{2-step}}(\phi')$.}
        \State{Optimize $\phi'$ with an gradient-based optimizer using $\nabla_{\phi'}\mathcal{L}_{\text{combined}}$.}

    \Until{$\mathcal{L}_{\text{combined}}$ converges.}
    
    \State{\textbf{Return} one-step model $v_{\phi'}$.}
    
    \end{algorithmic}
\end{algorithm}

\section{Experiments}
\label{sec:expr}

In this section, we provide empirical evaluation of~\method{} and compare it with prior arts.
The source code is available at \href{https://github.com/yuanzhi-zhu/SlimFlow}{https://github.com/yuanzhi-zhu/SlimFlow}. 


\subsection{Experimental Setup}
\label{sec:expr_setup}

\noindent\textbf{Datasets and Pre-trained Teacher Models.}~~
Our experiments are performed on the CIFAR10 \cite{krizhevsky2009learning} 32$\times$32 and the FFHQ \cite{karras2019style} 64$\times$64 datasets to demonstrate the effectiveness of \method{}. Additionally, we evaluated our method's performance on the ImageNet \cite{deng2009imagenet} 64$\times$64 dataset, focusing on conditional generation tasks.
The pre-trained large teacher models are adopted from the official checkpoints from previous works, Rectified Flow~\cite{liu2022rectified} and EDM~\cite{karras2022elucidating}.

\noindent\textbf{Implementation Details.}~~ For experiments on CIFAR10 32$\times$32 and the FFHQ 64$\times$64, we apply the U-Net architecture of NCSN$++$ proposed in \cite{song2020score}. For experiments on ImageNet 64$\times$64, we adopt a different U-Net architecture proposed in~\cite{dhariwal2021diffusion}.
In the CIFAR10 experiments, we executed two sets of experiments with two different teacher models:
one is the pre-trained 1-Rectitied Flow~\cite{liu2022flow} and the other is the pre-trained EDM model~\cite{karras2022elucidating}. 
For all the other experiments, we adopt only the pre-trained EDM model as the large teacher model, unless specifically noted otherwise.
All the experiments are conducted on 4 NVIDIA 3090 GPUs.
In Distillation, we found replacing $v_\phi(\mathbf{x}_1, 1)$ with $v_{\phi'}(\mathbf{x}_1, 1)$ in~\cref{eq:2-step} leads to high empirical performance as $v_{\phi'}$ is a better one-step generator and speed up the training by saving one forward of the 2-rectified flow, so we keep that in our practice.
More details can be found in the Appendix.

\noindent\textbf{Evaluation Metrics.}~~
 We use the Fréchet inception distance (FID)\cite{heusel2017gans} to evaluate the quality of generated images. 
 In our experiments, we calculate the FID by comparing 50,000 generated images with the training dataset using Clean-FID \cite{parmar2022aliased}.
We also report the number of parameters (\#Params), Multiply-Add Accumulation (MACs), and FLoating-point OPerations per second (FLOPs) as metrics to compare the computational efficiency of different models.
It is important to note that in this paper, both MACs and FLOPs refer specifically to the computation required for a single forward inference pass through the deep neural network.
We use Number of Function Evaluations (NFEs) to denote the number of inference steps.


\begin{table*}[!t]
\centering
\footnotesize
\resizebox{0.95\linewidth}{!}{
\begin{tabular}{l|l|cccc}
\toprule
Category & Method & \#Params \quad & NFE ($\downarrow$) \quad & FID ($\downarrow$) \quad & MACs ($\downarrow$) \\
\midrule 
\multirow{2}{*}{Teacher Model} & EDM \cite{karras2022elucidating} & 55.7M & 35 & 1.96 & 20.6G \\
& 1-Rectified Flow~\cite{liu2022flow} & 61.8M & 127 & 2.58 & 10.3G \\
\midrule 
\multirow{4}{*}{\shortstack[l]{{Diffusion}\\ {+ Pruning}}} 
& Proxy Pruning \cite{li2023not}
& 38.7M & 100 & 4.21 \\
& Diff-Pruning \cite{fang2024structural}
& 27.5M & 100 & 4.62 & 5.1G \\
& Diff-Pruning \cite{fang2024structural}
& 19.8M & 100 & 5.29 & 3.4G\\
& Diff-Pruning \cite{fang2024structural}
& 14.3M & 100 & 6.36 & 2.7G\\
\midrule 
\multirow{6}{*}{\shortstack[l]{{Diffusion}\\ {+ Fast Samplers}}} 
& AMED-Solver \cite{zhou2023fast} & 55.7M & 5 & 7.14 & 20.6G\\
& GENIE \cite{dockhorn2022genie}
& 61.8M & 10 & 5.28 & 10.3G\\
& 3-DEIS \cite{zhang2022fast}
& 61.8M & 10 & 4.17 & 10.3G\\
& DDIM \cite{song2020denoising}
& 35.7M & 10 & 13.36 & 6.1G\\
& DPM-Solver-2 \cite{lu2022dpm}
& 35.7M & 10 & 5.94 & 6.1G\\
& DPM-Solver-Fast \cite{lu2022dpm}
& 35.7M & 10 & 4.70 & 6.1G \\
\midrule 
\multirow{17}{*}{\shortstack[l]{{Diffusion}\\ {+ Distillation}}} 
& DSNO \cite{zheng2023fast}
& 65.8M & 1 & 3.78 & \\
& Progressive Distillation \cite{salimans2022progressive} & 60.0M & 1 & 9.12 & \\
& Score Mismatching~\cite{ye2023score}$^\dagger$ & 24.7M & 1 & 8.10 & \\
& TRACT \cite{berthelot2023tract} & 19.4M & 1 & 6.47 & \\
& Diff-Instruct~\cite{luo2023diff}$^\dagger$ & 55.7M & 1 & 4.53 & 20.6G \\
& TRACT \cite{berthelot2023tract} & 55.7M & 1 & 3.78 & 20.6G \\
& DMD \cite{yin2024one}$^\dagger$ & 55.7M & 1 & 3.77 & 20.6G \\
& Consistency Distillation (EDM teacher) \cite{song2023consistency} & 55.7M & 1 & {3.55}  & 20.6G \\
& 1-Rectified Flow (+distill) \cite{liu2022rectified}
 & 61.8M & 1 & 6.18 & 10.3G \\
& 2-Rectified Flow (+distill) \cite{liu2022rectified}
 & 61.8M & 1 & 4.85 & 10.3G\\
& 3-Rectified Flow (+distill) \cite{liu2022rectified}
 & 61.8M & 1 & 5.21 & 10.3G \\
& Consistency Distillation (EDM teacher) \cite{song2023consistency}$^*$ & 27.9M & 1 & 6.83 & 6.6G\\
& \textbf{\method{}} (EDM teacher) & 27.9M & 1 & 4.53 & 6.6G \\
& Knowledge Distillation~\cite{luhman2021knowledge}
 & 35.7M & 1 & 9.36 & 6.1G \\
& Consistency Distillation (EDM teacher) \cite{song2023consistency}$^*$ & 15.7M & 1 & 7.21 & 3.7G\\
& \textbf{\method{}} (EDM teacher) & 15.7M & 1 & 5.02 & 3.7G \\
& \textbf{\method{}} (1-Rectified Flow teacher) & 15.7M & 1 & 5.81 & 3.7G  \\
\bottomrule
\end{tabular}}
\caption{
\label{table:cifar10}
Comparison on CIFAR10.
`$*$' refers to reproduced results.
}
\end{table*}

\begin{figure}[!t]
\centering
\includegraphics[width=0.99 \linewidth]{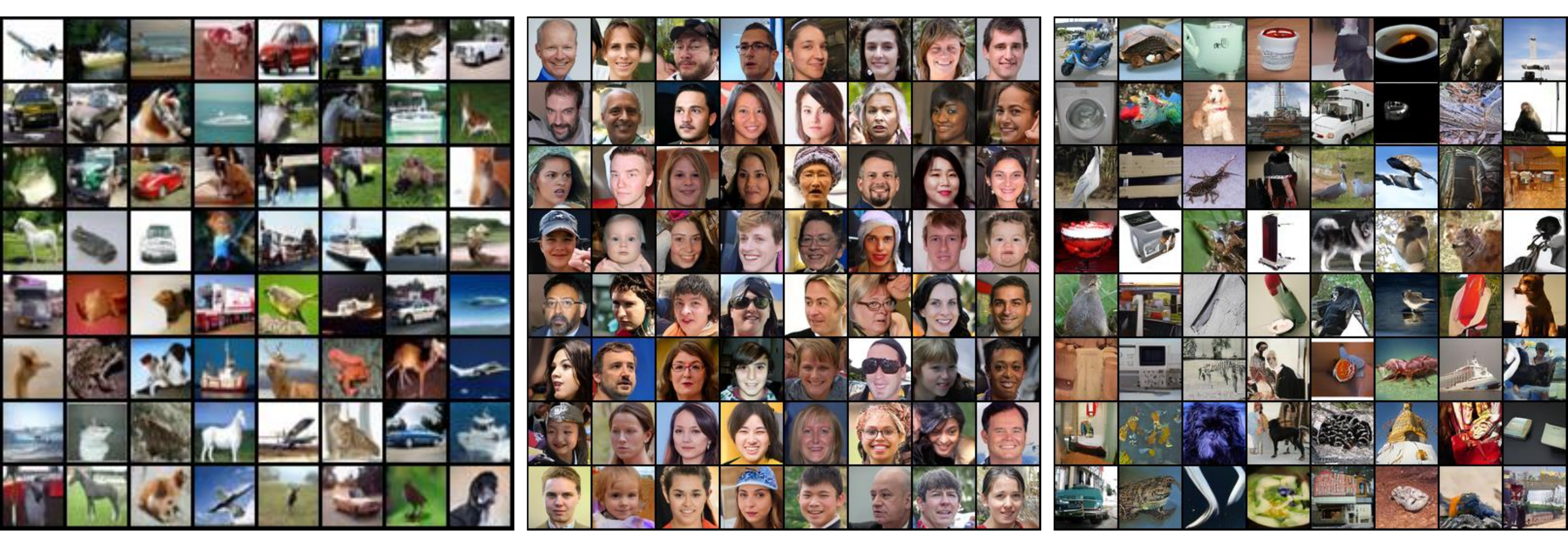}
\caption{Random generation from our best one-step small models on three different datasets. 
Left: CIFAR10 32$\times$32 (\#Params=27.9M). Mid: FFHQ 64$\times$64 (\#Params=27.9M). Right: ImageNet 64$\times$64 (\#Params=80.7M).}
\label{fig:one_step}
\end{figure}

\begin{table*}[!t]
\centering
\footnotesize
\resizebox{0.95\linewidth}{!}{
\begin{tabular}{l|l|ccccc}
\toprule
Dataset & Method & \#Params \quad & NFE ($\downarrow$) \quad & FID ($\downarrow$) \quad & MACs ($\downarrow$) \quad & FLOPs ($\downarrow$) \\ 
\midrule 
\multirow{5}{*}{\shortstack[l]{{FFHQ}\\ {64$\times$64}}}
& EDM \cite{karras2022elucidating} & 55.7M & 79 & 2.47 & 82.7G & 167.9G\\
& DDIM \cite{song2020denoising}
& 55.7M & 10 & 18.30 & 82.7G & 167.9G\\
& AMED-Solver \cite{zhou2023fast} & 55.7M & 5 & 12.54 & 82.7G & 167.9G \\
& BOOT \cite{gu2023boot} & 66.9M & 1 & 9.00 & 25.3G & 52.1G \\
& \textbf{\method{}} (EDM teacher) & 27.9M & 1 & 7.21 & 26.3G & 53.8G \\
& \textbf{\method{}} (EDM teacher) & 15.7M & 1 & 7.70 & 14.8G & 30.4G \\
\midrule
\multirow{11}{*}{\shortstack[l]{{ImageNet}\\ {64$\times$64}}}
& EDM \cite{karras2022elucidating} & 295.9M & 79 & 2.37 & 103.4G & 219.4G \\
& DDIM \cite{song2020denoising}
& 295.9M & 10 & 16.72 & 103.4G & 219.4G \\
& AMED-Solver \cite{zhou2023fast} & 295.9M & 5 & 13.75 & 103.4G & 219.4G \\
& DSNO \cite{zheng2023fast} & 329.2M & 1 & 7.83 &  \\
& Progressive Distillation \cite{salimans2022progressive} & 295.9M & 1 & 15.39 & 103.4G & 219.4G \\
& Diff-Instruct~\cite{luo2023diff} & 295.9M & 1 & 5.57 & 103.4G & 219.4G \\
& TRACT \cite{berthelot2023tract} & 295.9M & 1 & 7.43 & 103.4G & 219.4G \\
& DMD \cite{yin2024one} & 295.9M & 1 & 2.62 & 103.4G & 219.4G \\
& Consistency Distillation \cite{song2023consistency} & 295.9M & 1 & 6.20 & 103.4G & 219.4G \\
& Consistency Training \cite{song2023consistency} & 295.9M & 1 & 13.00 & 103.4G & 219.4G \\
& BOOT \cite{gu2023boot} & 226.5M & 1 & 16.30 & 78.2G & 157.4G \\
& \textbf{\method{}} (EDM teacher) & 80.7M & 1 & 12.34 & 31.0G & 67.8G \\
\bottomrule
\end{tabular}}
\caption{
\label{table:FFHQ_ImageNet}
Comparison on FFHQ 64$\times$64 and ImageNet 64$\times$64. 
}
\end{table*}

\subsection{Empirical Results}
\begin{figure}[!t]
\centering
\begin{subfigure}[b]{0.5\textwidth}
\centering
\begin{overpic}[width=0.9 \linewidth]{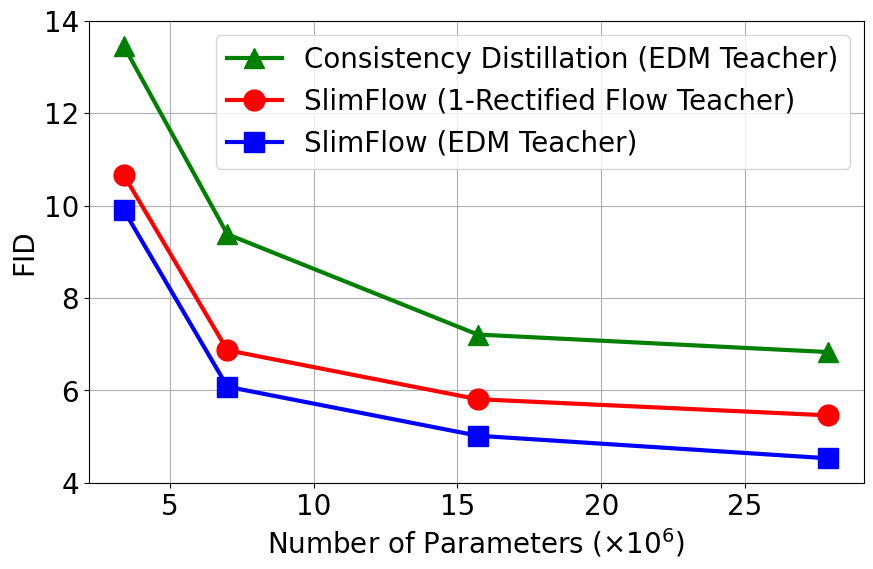}
\put(-3,60.){\color{black}{(a)}}
\end{overpic}
\phantomcaption
\label{fig:fid_a}
\end{subfigure}%
\begin{subfigure}[b]{0.5\textwidth}
\centering
\begin{overpic}[width=0.9 \linewidth]{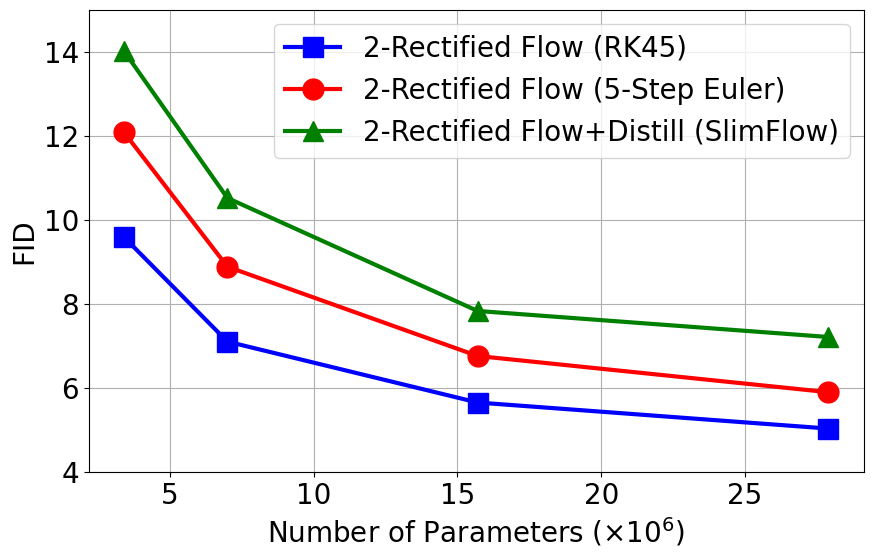}
\put(-3.5,58.5){\color{black}{(b)}}
\end{overpic}
\phantomcaption
\label{fig:fid_b}
\end{subfigure}
\caption{
(a) Comparison of models trained with different methods on CIFAR10. 
(b) Comparison between 2-rectified flow and the distilled one-step generator on CIFAR10.}
\label{fig:fid}
\end{figure}

\noindent\textbf{\method{} on CIFAR10.}~~
We report the results on CIFAR10 in Table~\ref{table:cifar10}.
For all the \method{} models, we train them using \anflow{} with the data pairs generated from the large teacher model for 800,000 iterations, then distilling them to one-step generator with flow-guided distillation for 400,000 iterations. 
For the consistency distillation baselines, we train them for 1,200,000 iterations with the same EDM teacher to ensure a fair comparison.
With only 27.9M parameters, one-step \method{} outperforms the 61.8M 2-Rectified Flow+Distill model (FID: $4.53 \leftrightarrow 4.85$). With the less than 20M parameters, \method{} gives better FID (5.02, \#Params=15.7M) than TRACT (6.48, \#Params=19.4M) and Consistency Distillation (7.21, \#Params=15.7M). Additional SlimFlow results with different parameters can be found in \cref{fig:fid_a} for comparison with Consistency Distillation \cite{song2023consistency} and in \cref{fig:fid_b} for comparison with 2-rectified flow.

\noindent\textbf{\method{} on FFHQ and ImageNet.}~~
We report the results of \method{} on FFHQ 64$\times$64 and ImageNet 64$\times$64 in Table~\ref{table:FFHQ_ImageNet}. For ImageNet 64$\times$64, the models are trained in the conditional generation scenarios where class labels are provided.
In FFHQ, our \method{} models surpasses BOOT with only 15.7M parameters. In ImageNet, \method{} obtains comparable performance as the 295.9M models trained with Consistency Training, BOOT and Progressive Distillation, using only 80.7M parameters. These results demonstrates the effectiveness of \method{} in training efficient one-step generative models.

\begin{figure}[!t]
\centering
\begin{overpic}[width=0.95 \linewidth]{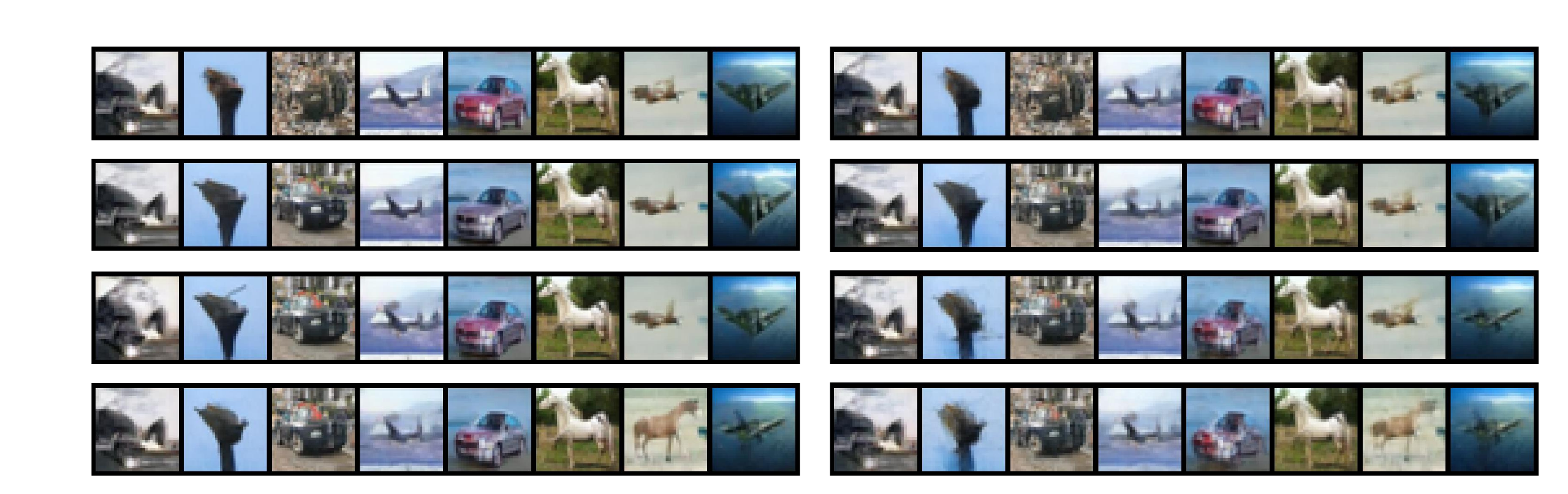}
\put(19,29.3){\color{black}\scriptsize{Multi-Step RK45}}
\put(67,29.3){\color{black}\scriptsize{One-Step Euler}}
\put(-1.5,25){\color{black}\scriptsize{27.9M}}
\put(-1.5,17.6){\color{black}\scriptsize{15.7M}}
\put(-1.5,10.5){\color{black}\scriptsize{7.0M}}
\put(-1.5,3.2){\color{black}\scriptsize{3.4M}}
\end{overpic}
\caption{CIFAR10 samples from 2-rectified flow models trained with \anflow{}. All images are generated with the same set of random noises. }
\label{fig:2flow_cifar}
\end{figure}
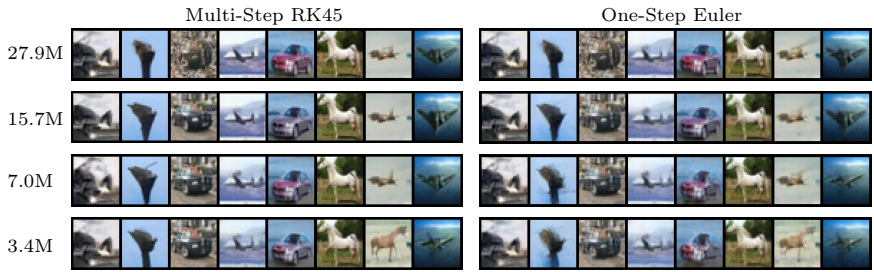

\begin{figure}[!t]
\centering
\begin{subfigure}[b]{0.48\textwidth}
\centering
\begin{overpic}[width=0.9 \linewidth]{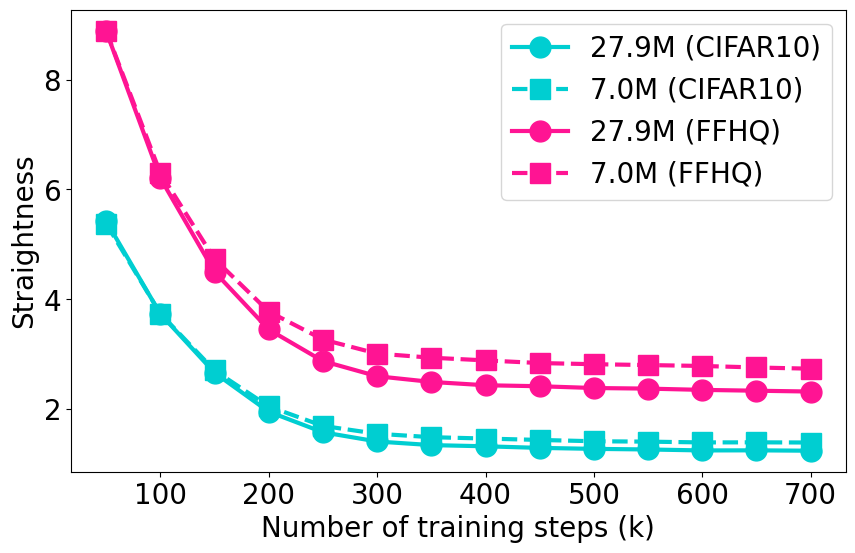}
\put(-3.5,60.){\color{black}{(a)}}
\end{overpic}
\phantomcaption
\label{fig:straightness_a}
\end{subfigure}%
\begin{subfigure}[b]{0.5\textwidth}
\centering
\begin{overpic}[width=0.9 \linewidth]{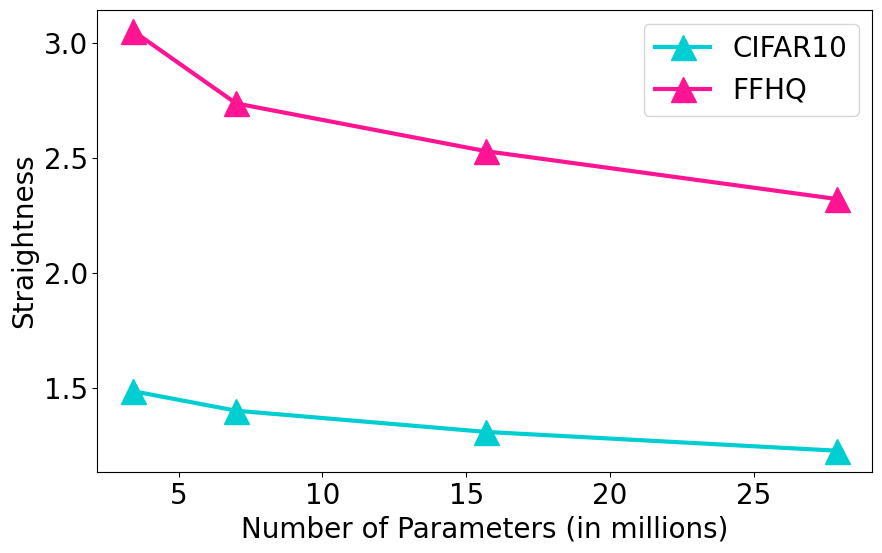}
\put(-3.5,57.5){\color{black}{(b)}}
\end{overpic}
\phantomcaption
\label{fig:straightness_b}
\end{subfigure}
\caption{(a) Straightness of 2-rectified flows with different sizes during \anflow{}. (b) Final straightness of 2-rectified flows with different numbers of parameters. }
\label{fig:straightness}
\end{figure}

\noindent\textbf{Analysis of Annealing Reflow.}~~
We examine the straightening effect of \anflow{}.
We measured the straightness in the \anflow{} stage of models with different sizes on both the CIFAR10 32$\times$32 and the FFHQ 64$\times$64 dataset. 
Here, straightness is defined as in 
\cref{eq:straightness} following~\cite{liu2022flow, liu2023instaflow}.
In \cref{fig:straightness_a}, straightness decreases as $\beta(k)$ gradually approaches 0.
In \cref{fig:straightness_b}, we observe that the straightness of the resulting 2-rectified flows decreases as their number of parameters increase.
In \cref{fig:2flow_cifar}, samples generated with four 2-rectified flows with different samplers are presented.
These results demonstrate the effectiveness of \anflow{} in learning straight generative flows.

\begin{table*}[!t]
\centering
\footnotesize
\begin{minipage}{0.38\linewidth}
  \begin{center}
  \setlength\tabcolsep{5pt}
  \begin{tabular}{ccc}
    \toprule
    Annealing & H-Flip & FID ($\downarrow$) \\
    \midrule 
     - & - & 5.84\\
     - & \checkmark & 5.06\\
    \checkmark & - & 5.46\\
    \checkmark & \checkmark & 4.51\\
    \bottomrule
  \end{tabular}
  \end{center}
\caption{Ablation study on the reflow strategy of \method{}.}
\label{tab:anneal_1}
\end{minipage}
\hfill
\begin{minipage}{0.58\linewidth}
  \begin{center}
    \begin{tabular}{cc}
    \toprule
    $\beta(k)$ & FID ($\downarrow$) \\
    \midrule 
    $0$ & 5.06\\
    $\exp(-k / K_{step})$ & 4.78\\
    $\cos(\pi \min(1, k / 2K_{step})/2)$ & 4.79\\
    $(1+\cos(\pi \min(1,k/2K_{step})))/2$ & 4.70\\
    $1 - \min(1, k / 6 K_{step})$ & 4.51\\
    \bottomrule
    \end{tabular}
    \caption{
    \label{tab:anneal_2}
    Ablation study on the annealing strategy of \method{}. 
    Recall that $k$ is the number of training iterations. We set $K_{step}=50,000$.
    }
  \end{center}
\end{minipage}
\end{table*}

\subsection{Ablation Study}
\noindent\textbf{\anflow{}.}~~
We examine the design choices in \anflow{}. We train 2-rectified flow for 800,000 iterations and measure its FID with RK45 solver. 
In Table~\ref{tab:anneal_1}, we report the influence of the \anflow{} strategy and the effectiveness of exploiting the intrinsic symmetry of reflow. It can be observed that both the annealing strategy and the intrinsic symmetry improve the performance of 2-rectified flow, and their combination gives the best result. In Table~\ref{tab:anneal_2}, we analyze the schedule of $\beta(k)$. When $\beta(k)=0$, it is equivalent to training 2-rectified flow with random initialization. All other schedules output lower FID than $\beta(k)=0$, showing the usefulness of our annealing strategy. We adopt $\beta(k)=1-\min(1, k / 6 K_{step})$ as our default schedule in our experiments.

\noindent\textbf{Flow-Guided Distillation.}~~In Table~\ref{table:distillation}, we analyze the influence of different choices in our distillation stage. We found that using our 2-step regularization boosts the FID the the distilled one-step model. Using $\ell_2$ loss, it improves the FID from 7.90 to 7.30. Using LPIPS loss, it improves the FID from 6.43 to 5.81. 
We use the model architecture with 15.7M parameters for all ablation studies.

\begin{table*}[!t]
\centering
\setlength\tabcolsep{7pt}
\small{
\begin{tabular}{ccccc}
\toprule
Initialize from $v_\phi$ \qquad & Source of $\mathcal{D}_{distill}$ \qquad & Loss $\mathbb{D}$ \qquad & with $\mathcal{L}_{\text{2-step}}$ \qquad & FID \\ \midrule 
-        & 1-rectified flow $v_\theta$    & $\ell_2$  & -          & 12.09   \\ 
\checkmark        & 1-rectified flow $v_\theta$    & $\ell_2$    & -          & 8.98    \\
-        & 2-rectified flow $v_\phi$     & $\ell_2$   & -          & 9.19    \\ 
\checkmark        & 2-rectified flow $v_\phi$    & $\ell_2$   & -          & 7.90    \\ 
\checkmark        & 2-rectified flow $v_\phi$    & $\ell_2$    & \checkmark          & 7.30    \\
\checkmark        & 2-rectified flow $v_\phi$    & LPIPS        & -          & 6.43    \\
\checkmark        & 2-rectified flow $v_\phi$    & LPIPS        & \checkmark          & 5.81    \\ 
\bottomrule
\end{tabular}
}
\caption{Ablation study on the distillation stage of \method{} on CIFAR10. }
\label{table:distillation}
\end{table*}

\section{Related Work}
\noindent\textbf{Reducing the Inference Steps of Diffusion Models.}~~ 
Diffusion models generate new samples through an iterative process, where a noisy image is repeatedly denoised by a neural network. To accelerate this process, researchers have proposed various methods, which can be categorized into two main approaches.

The first category is training-free methods, which aim to reduce the number of inference steps for existing diffusion models by optimizing the sampling process. By reformulating diffusion models into PF-ODEs \cite{song2020score}, numerous techniques are introduced to accelerate sampling while minimizing quality loss.
Some of the notable examples in this area include DDIM \cite{song2020denoising}, EDM \cite{karras2022elucidating}, DEIS \cite{zhang2022fast} and DPM-solver \cite{lu2022dpm}.
A recent advancement in this category is the AMED-Solver~\cite{zhou2023fast}, which leverages the mean value theorem to minimize discretization error, achieving high-quality generation with even fewer function evaluations.

Beyond these fast ODE salvers, distillation-based methods aim to achieve few-step sampling, or even one-step sampling, by training a new student model with the pre-trained multi-step diffusion models as teachers. 
Progressive Distillation~\cite{salimans2022progressive} proposed to repeatedly train a student network whose step size is twice as the step size of the teacher and set the student as a new teacher for the next round.
BOOT~\cite{gu2023boot} suggested distilling the knowledge of the teacher models in a data-free manner with the help of the signal-ODE.
Distribution Matching Distillation~\cite{yin2024one} extended the idea of Variational Score Distillation~\cite{wang2023prolificdreamer} to train a one-step generator by alternatively updating the one-step generator and a fake data score function.
Consistency models~\cite{song2023consistency, luo2023latent} are a new family of generative models that trains few-step diffusion models by applying consistency loss.
Additionally, several methods have incorporated adversarial training to enhance the performance of one-step diffusion models~\cite{ye2023score,kim2023consistency,xu2023ufogen}.

\noindent\textbf{Reducing the Size of Diffusion Models.}~~
The increasing demand for low-budget and on-device applications necessitates the development of compact diffusion models, as current state-of-the-art models like Stable Diffusion are typically very large. To address this challenge, researchers have explored various compression techniques, primarily focusing on network pruning and quantization methods.

Network pruning involves selectively removing weights from the network and subsequently fine-tuning the pruned model to maintain performance comparable to the pre-trained version.
Diff-Pruning \cite{fang2024structural} utilizes Taylor expansion over pruned timesteps to identify non-contributory diffusion steps and important weights through informative gradients.
BK-SDM \cite{kim2023bk} discovers that block pruning combined with feature distillation is an efficient and sufficient strategy for obtaining lightweight models.

Another line of work is model quantization, 
which aims to reduce the storage and computational requirements of diffusion models during deployment \cite{wang2023towards, li2023q, he2023efficientdm, shang2023post, li2024q, huang2023tfmq}.
Q-diffusion \cite{li2023q} introduces timestep-aware calibration and split short-cut quantization, tailoring post-training quantization (PTQ) methods specifically for diffusion models.
EfficientDM \cite{he2023efficientdm} proposes a quantization-aware variant of the low-rank adapter (QALoRA), achieving Quantization-Aware Training (QAT)-level performance with PTQ-like efficiency.


%

In summary, \method{} advances the state of the art by addressing the dual challenge of minimizing both inference steps and neural network size, with the ultimate goal of developing the most efficient diffusion models.
While our approach shares some similarities with MobileDiffusion \cite{zhao2023mobilediffusion}, which also explores efficient structure design and acceleration of diffusion models, \method{} distinguishes itself in several key aspects:
(1) \method{} is built upon the rectified flow framework, which provides more stable training dynamics than the GAN-based training used in MobileDiffusion; 
(2) Unlike MobileDiffusion's focus on fine-grained specific network structure optimization, \method{} offers a general framework applicable to a wide range of efficient network architectures. This flexibility allows our approach to leverage advancements in efficient network design across the field, including MobileDiffusion's efficient text-to-image network.

\section{Limitations and Future Works}
While our approach demonstrates advancements in efficient one-step diffusion models, we acknowledge several limitations and areas for future research. The quality of our one-step model is inherently bounded by the capabilities of the teacher models, specifically the quality of synthetic data pairs they generate. This limitation suggests a direct relationship between teacher model performance and the potential of our approach. In the future, we will leverage more advanced teacher models and real datasets. Besides, we plan to extend our method to more network architectures, e.g., transformers and pruned networks. Finally, given sufficient computational resources, we aim to apply our approach to more complex diffusion models, such as Stable Diffusion.

\section{Conclusions}
\label{sec:conslusion}


In this paper, we introduced \method{}, an innovative approach to developing efficient one-step diffusion models. 
Our method can significantly reduce model complexity while preserving the quality of one-step image generation, as evidenced by our results on CIFAR-10, FFHQ, and ImageNet datasets. 
This work paves the way for faster and more resource-efficient generative modeling, broadening the potential for real-world applications of diffusion models.


\par\vfill\par

\newpage
\section*{Acknowledgements}
We would like to thank Zewen Shen for his insightful discussion during this work.

%
%
\bibliographystyle{splncs04}
\bibliography{egbib}
\clearpage

\setcounter{section}{0}
\renewcommand{\thesection}{\Alph{section}}

\section*{Appendix}

\section{Results of \anflow{} with Different Fixed $\beta$}
\label{append:beta_flow}
\begin{figure}[ht]
\centering
\begin{overpic}[width=0.99 \linewidth]{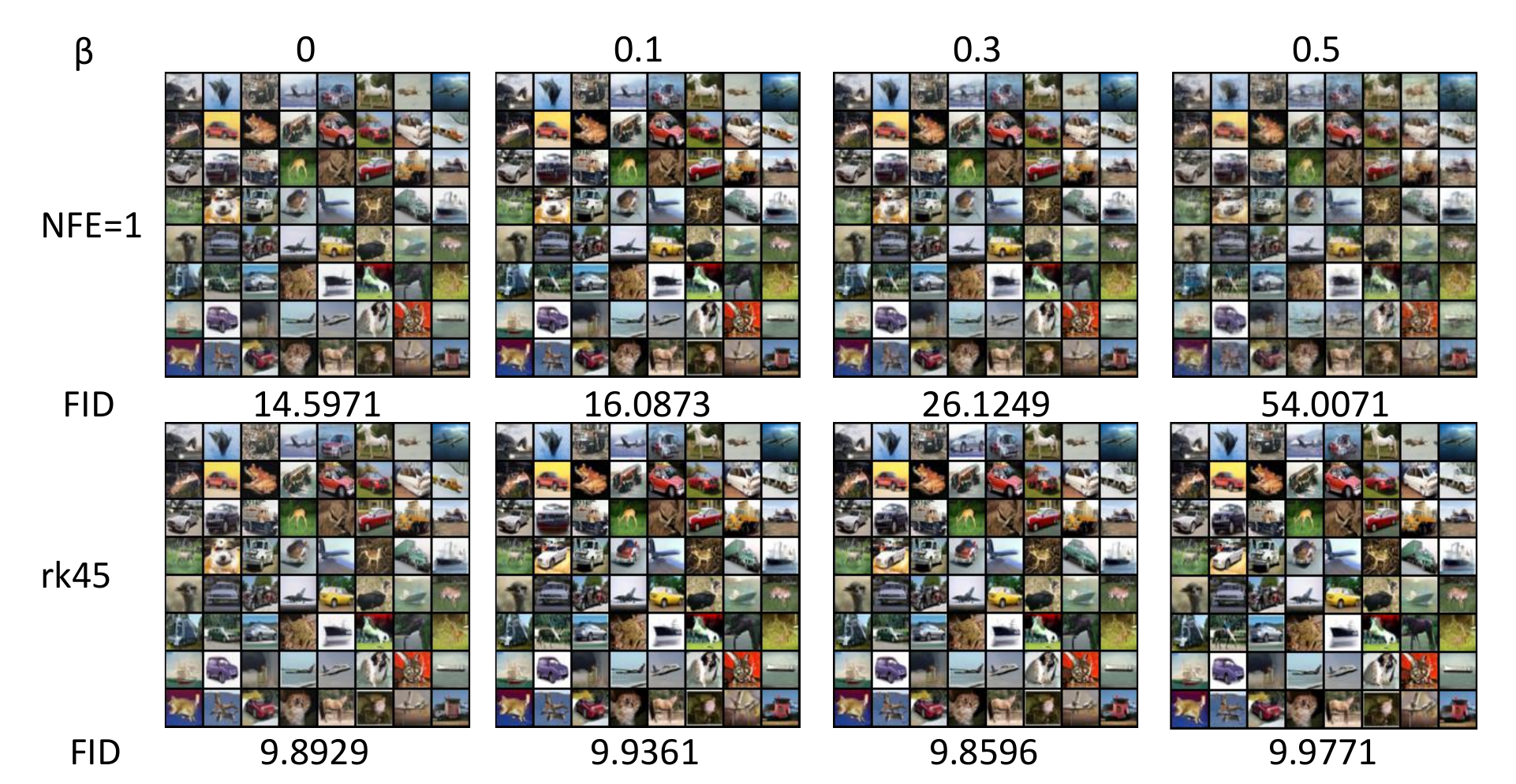}
\end{overpic}
\caption{Result on models with 15M parameters trained with different fixed $\beta$ values. The FIDs are calculated with only 5k generated samples.}
\label{fig:fix_beta}
\end{figure}

We present the sampled images and the corresponding FID over 5k images of \anflow{} training of different fixed $\beta$ using the same data pairs in \cref{fig:fix_beta}.
We test the results with four different $\beta$ values: \{0, 0.1, 0.3, 0.5\} and found that the RK45 samples are similar in four different models but the quality of one-step generated samples is much worse when $\beta$ is large. This observation validates the marginal preserving property of the \anflow{} training.

\section{More Experimental Details}
\label{append:expr}

\begin{table*}[!t]
\centering
\footnotesize
\resizebox{0.99\linewidth}{!}{
\begin{tabular}{lccccccc}
\toprule
 & A & B & C & D$^{\dagger}$ & E & F & G \\
\midrule 
\#Blocks & 4 & 4 & 2 & 2 & 2 & 1 & 1 \\
Base Channels & 128 & 128 & 128 & 96 & 64 & 64 & 64\\
Channel Multiplier & (1, 2, 2, 2) & (2, 2, 2) & (1, 2, 2) & (1, 2, 2) & (1, 2, 2) & (1, 1, 2) & (1, 1)\\
Attention resolutions & (16,) & (16,) & (16,) & (16,) & (16,) & (16,) & (16,) \\
\midrule 
Batch Size & - & - & 128 & 128 & 128 & 128 & 128 \\
Batch Size (FFHQ distillation) & - & - & 64 & 64 & 128 & 128 & - \\
\midrule 
\#Paras & 61.8M & 55.7M & 27.9M & 15.7M & 7.0M & 3.4M & 1.2M\\
\midrule 
MACs & 10.3G & 20.7G & 6.6G & 3.7G & 1.6G & 0.7G & 0.5G\\
FLOPs & 22.0G & 42.7G & 13.9G & 7.9G & 3.6G & 1.5G & 1.2G\\
\bottomrule
\end{tabular}}
\vspace{0.3cm}
\caption{
\label{table:architecture}
Architecture configurations that are used in this work for CIFAR10 and FFHQ. $\dagger$ represents the default configuration for ablations. MACs and FLOPs are calculated with input shape $(1, 3, 32, 32)$.
}
\end{table*}

\begin{table*}[!t]
\centering
\footnotesize
\resizebox{0.8\linewidth}{!}{
\begin{tabular}{llccc}
\toprule
& & H & I & J \\
\midrule 
\multirow{4}{*}{\shortstack[l]{{Architecture}\\ {Configuration}}} 
& \#Blocks & 3 & 2 & 2 \\
& Base Channels & 192 & 128 & 128 \\
& Channel Multiplier & (1, 2, 3, 4) & (1, 2, 2, 4) & (1, 2, 2, 2) \\
& Attention resolutions & (32,16,8) & (32, 16) & (32, 16) \\
\midrule 
\multirow{2}{*}{\shortstack[l]{{Training}}} 
& Batch Size & - & 96 & 96 \\
& Batch Size (distillation) & - & 64 & 64 \\
\midrule 
\multirow{3}{*}{\shortstack[l]{{Model Size}}} 
& \#Paras & 259.9M & 80.7M & 44.7M \\
& MACs & 103.4G & 31.0G & 28.1G \\
& FLOPs & 219.4G & 67.8G & 61.9G \\
\bottomrule
\end{tabular}}
\vspace{0.3cm}
\caption{
\label{table:architecture_imgnet}
Architecture configurations that are used in this work for the ImageNet dataset. MACs and FLOPs are calculated with input shape $(1, 3, 64, 64)$.
}
\end{table*}

In \cref{table:architecture}, we list all the architecture choices and related hyper-parameters in our experiments on CIFAR10 and FFHQ. In \cref{table:architecture_imgnet}, we list all the architecture choices and related hyper-parameters in our experiments on ImageNet.
The training of all the networks is smoothed by EMA with a ratio of 0.999999. Adam optimizer is adopted with a learning rate of $2e-4$ and the dropout rate is set to $0.15$, following \cite{liu2022rectified}.
For the \anflow{} training, we use $\ell_2$ loss with a uniform loss weight; for the distillation, we switch to the LPIPS loss.
Most of the ablation experiments are conducted with configuration $D$ using the data pairs from the 1-rectified flow teacher.

For the experiment on ImageNet, we also trained a model with configuration J in \cref{table:architecture_imgnet}, which has almost half the parameters but similar MACs to configuration I. 
We get an FID of 8.86 from the 2-rectified flow with RK45 sampler (NFE$\approx$40) and a final FID of 12.52 on the final one-step flow. The comparison is listed in \cref{table:imagenet_fid}.

\begin{table*}[!t]
\centering
\footnotesize
\resizebox{0.99\linewidth}{!}{
\begin{tabular}{ccccccc}
\toprule
 & Channel Multiplier & \#Paras & MACs & FLOPs & FID (2-rectified flows) & FID (distilled flows) \\
\midrule 
I & (1, 2, 2, 4) & 80.7M & 31.0G & 67.8G & 8.89 & 12.34\\
J & (1, 2, 2, 2) & 44.7M & 28.1G & 61.9G & 8.86 & 12.52\\
\bottomrule
\end{tabular}}
\vspace{0.3cm}
\caption{
\label{table:imagenet_fid}
Comparison between two ImageNet experiments in this work.
}
\end{table*}

For the \anflow{} training, we use 50k data pairs from the 1-rectified flow, 100k data pairs from EDM on CIFAR10 dataset, 200k data pairs from EDM on FFHQ dataset, and 400k data pairs from EDM on the ImageNet dataset. For distillation, we always simulate 500k data pairs from the 2-rectified flows.

For distillation, the loss after replacing the 2-rectified flow with the one-step model as mentioned in \cref{sec:expr_setup} is:
\begin{equation}
\label{eq:new_2-step}
    \mathcal{L}'_{\text{2-step}}(\phi'):= \mathbb{E}_{\mathbf{x}_1 \sim \pi_1} \left [ \int_0^1 \mathbb{D}(\mathbf{x}_1 - (1-t) \mathbf{v}_{\phi'}(\mathbf{x}_1, 1) - t\mathbf{v}_\phi(\mathbf{x}_t, t), \mathbf{x}_1 - \mathbf{v}_{\phi'}(\mathbf{x}_1, 1)) \text{d} t \right].
\end{equation}
where a stop-gradient operation is added to the first $\mathbf{v}_{\phi'}$. This will help the convergence of the training in practice and save one forward step of the 2-rectified flow.

All of the reference statistics for computing FID are from EDM \cite{karras2022elucidating}.
All of the straightness is calculated using 100 Euler steps and averaging over 256 images, with the following \cref{eq:straightness}.

\section{Additional Samples from \method{}}
\label{append:images}

In this section, we provide some additional samples from our one-step models.

\begin{figure}
\centering
\begin{overpic}[width=0.99 \linewidth]{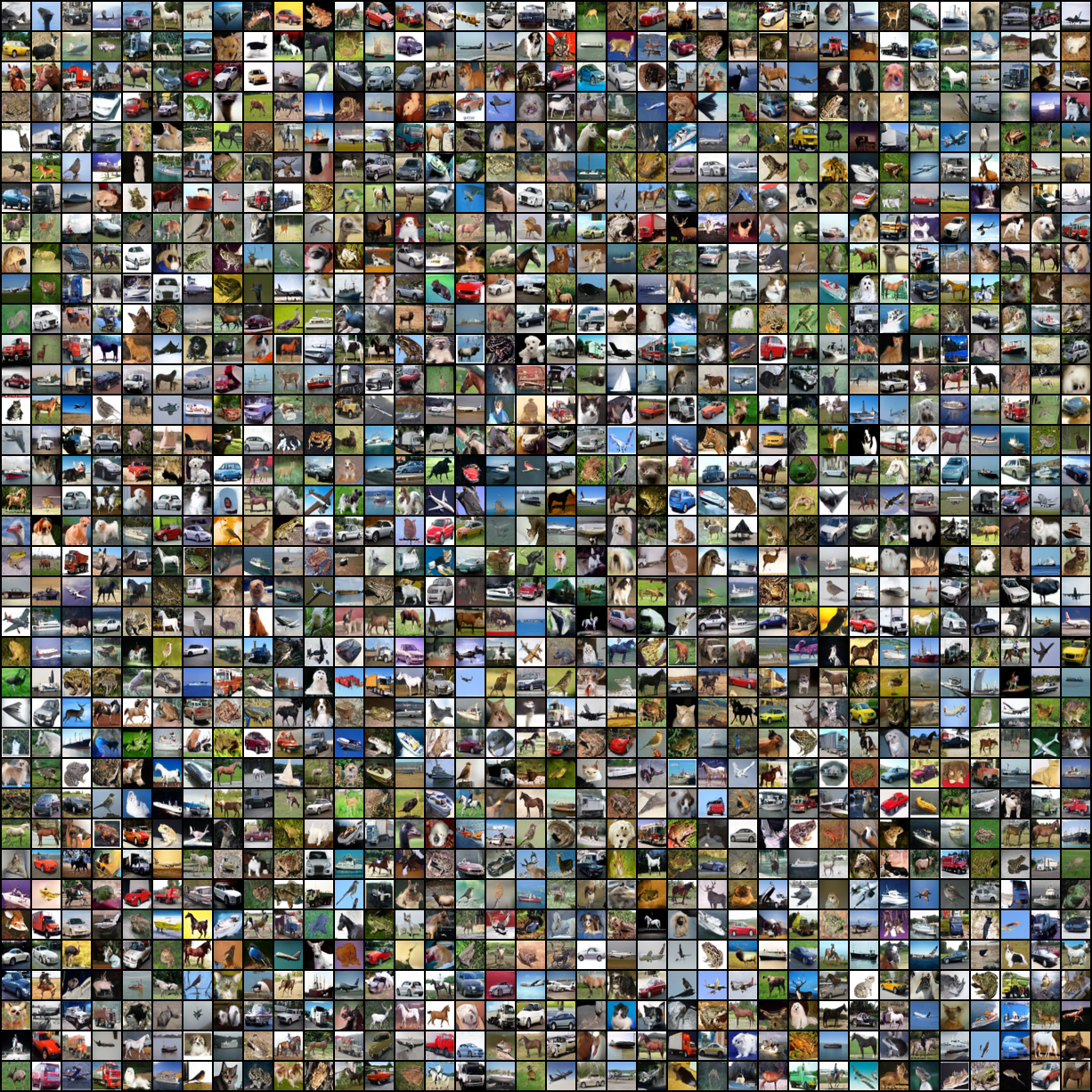}
\end{overpic}
\caption{Uncurated samples from unconditional CIFAR-10 32$\times$32 using \method{} with single step generation (FID=4.53).}
\label{fig:CIFAR_euler_sample_1}
\end{figure}

\begin{figure}
\centering
\begin{overpic}[width=0.99 \linewidth]{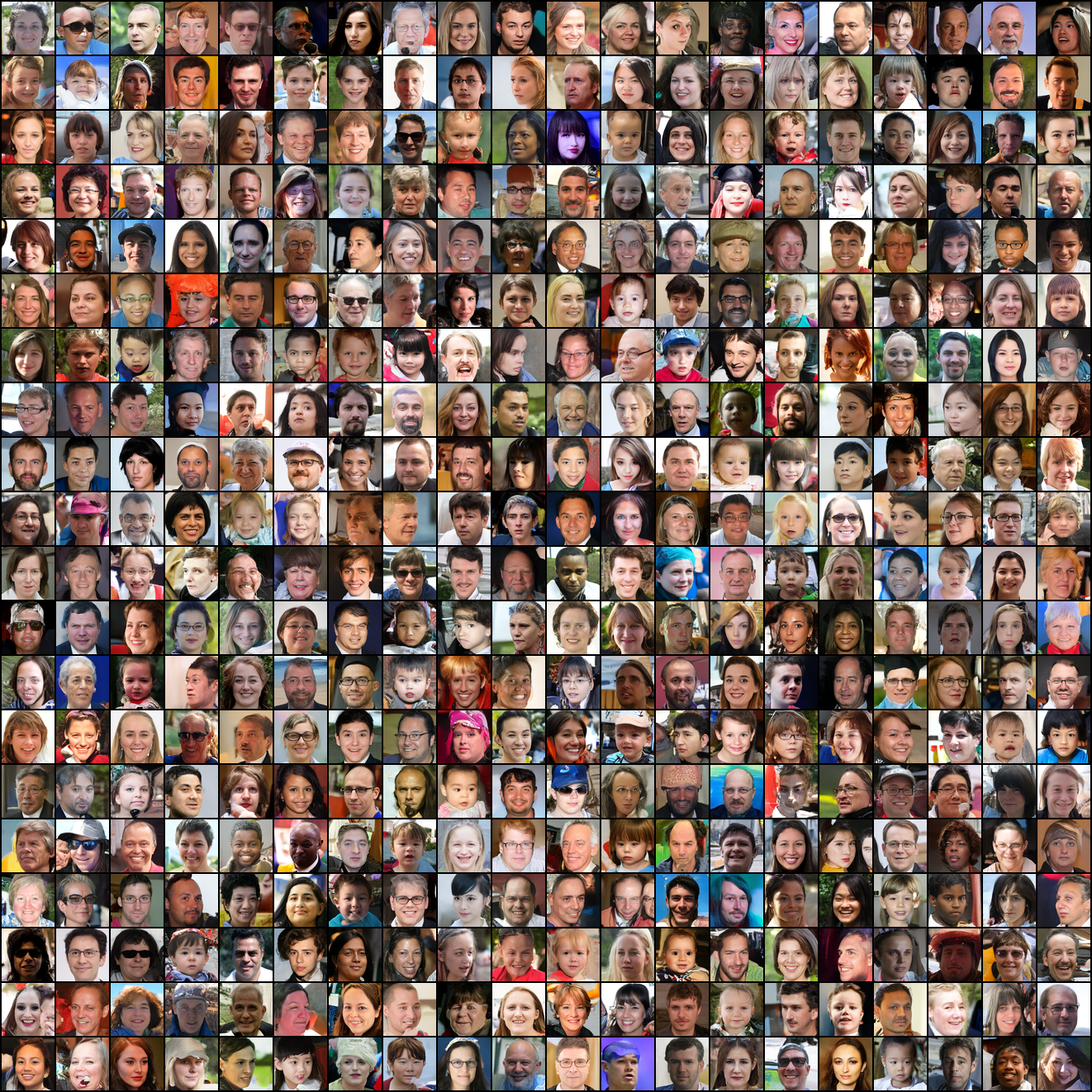}
\end{overpic}
\caption{Uncurated samples from unconditional FFHQ 64$\times$64 using \method{} with single step generation (FID=7.21).}
\label{fig:FFHQ_euler_sample_1}
\end{figure}

\begin{figure}
\centering
\begin{overpic}[width=0.99 \linewidth]{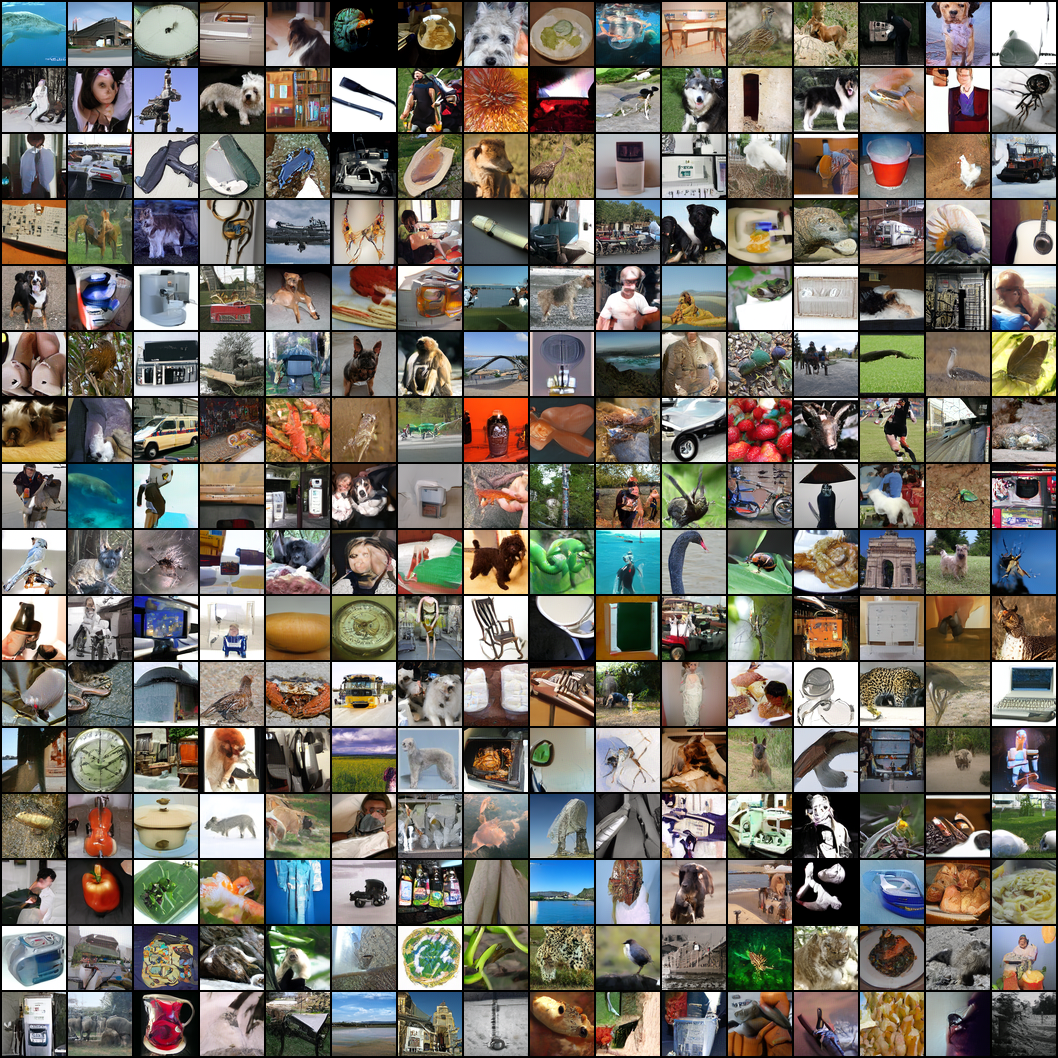}
\end{overpic}
\caption{Uncurated samples with random class labels from conditional ImageNet 64$\times$64 using \method{} with single step generation (FID=12.34).}
\label{fig:ImageNet_euler_sample_1}
\end{figure}

\begin{figure}
\centering
\begin{overpic}[width=0.99 \linewidth]{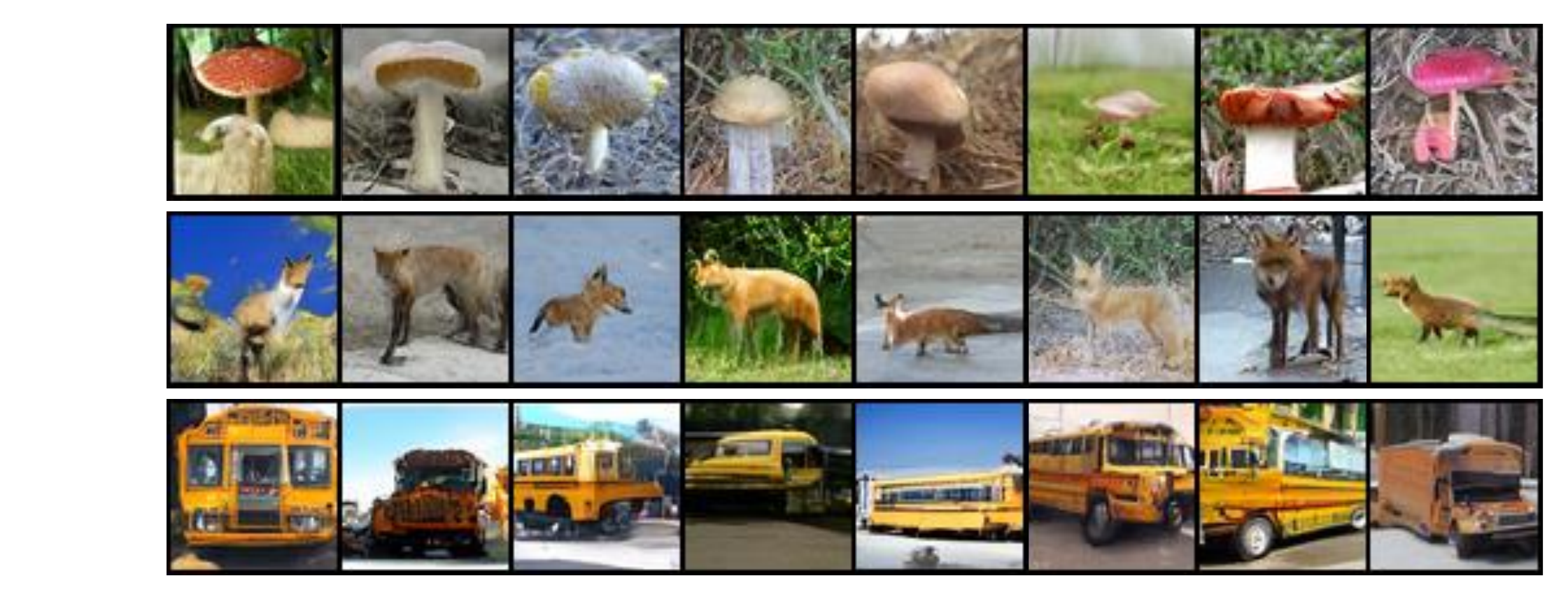}
\put(-1.5,30.5){\color{black}\scriptsize{Mushroom}}
\put(-1.5,18.5){\color{black}\scriptsize{Red fox}}
\put(-1.5,7){\color{black}\scriptsize{School bus}}
\end{overpic}
\caption{Uncurated samples with three given classes from conditional ImageNet 64$\times$64 using \method{} with single step generation (FID=12.34).}
\label{fig:imagenet}
\end{figure}

\end{document}